\pdfoutput=1

\documentclass[11pt]{article}

\usepackage[]{EMNLP2022}

\usepackage{times}
\usepackage{latexsym}

\usepackage[T1]{fontenc}

\usepackage[utf8]{inputenc}

\usepackage{microtype}
\usepackage{algorithm}
\usepackage{algorithmic}
\usepackage{booktabs}
\usepackage{amsmath}
\usepackage{amsfonts}
\usepackage{multirow}
\usepackage{bbding}
\usepackage{mathrsfs}
\usepackage{subfigure}
\usepackage{amssymb}
\usepackage{graphicx} 
\usepackage{amsthm}

\usepackage{inconsolata}

\title{Watch the Neighbors:  A Unified K-Nearest Neighbor Contrastive Learning Framework for OOD Intent Discovery}

\author{Yutao Mou$^{1*}$, Keqing He$^{2*}$, Pei Wang$^{1}$, Yanan Wu$^{1}$ \\
{\bf Jingang Wang$^{2}$,} {\bf Wei Wu$^{2}$,} {\bf Weiran Xu$^{1}$}\thanks{\ \ The first two authors contribute equally. Weiran Xu is the corresponding author.}\\
  $^1$Beijing University of Posts and Telecommunications, Beijing, China\\
$^{2}$Meituan, Beijing, China\\
  \texttt{\{myt,wangpei,yanan.wu,xuweiran\}@bupt.edu.cn}\\
  \texttt{\{hekeqing,wangjingang,wuwei\}@meituan.com}
}

\begin{document}
\maketitle
\begin{abstract}
Discovering out-of-domain (OOD) intent is important for developing new skills in task-oriented dialogue systems. The key challenges lie in how to transfer prior in-domain (IND) knowledge to OOD clustering, as well as jointly learn OOD representations and cluster assignments. Previous methods suffer from in-domain overfitting problem, and there is a natural gap between representation learning and clustering objectives. In this paper, we propose a unified K-nearest neighbor contrastive learning framework to discover OOD intents. Specifically, for IND pre-training stage, we propose a KCL objective to learn inter-class discriminative features, while maintaining intra-class diversity, which alleviates the in-domain overfitting problem. For OOD clustering stage, we propose a KCC method to form compact clusters by mining true hard negative samples, which bridges the gap between clustering and representation learning. Extensive experiments on three benchmark datasets show that our method achieves substantial improvements over the state-of-the-art methods. \footnote{We release our code at \url{https://github.com/myt517/KCOD}}

\end{abstract}

\section{Introduction}




Out-of-domain (OOD) intent discovery aims to group new unknown intents into different clusters, which helps identify potential development directions and develop new skills in a task-oriented dialogue system \cite{Lin2020DiscoveringNI, Zhang2021DiscoveringNI, mou-etal-2022-disentangled}. Different from traditional text clustering task, OOD discovery considers how to leverage the prior knowledge of known in-domain (IND) intents to enhance discovering unknown OOD intents, which makes it challenging to directly apply existing clustering algorithms \cite{MacQueen1967SomeMF, Xie2016UnsupervisedDE, Chang2017DeepAI, Caron2018DeepCF} to the OOD discovery task.


\begin{figure}[t]
    \centering
    \resizebox{.485\textwidth}{!}{
    \includegraphics{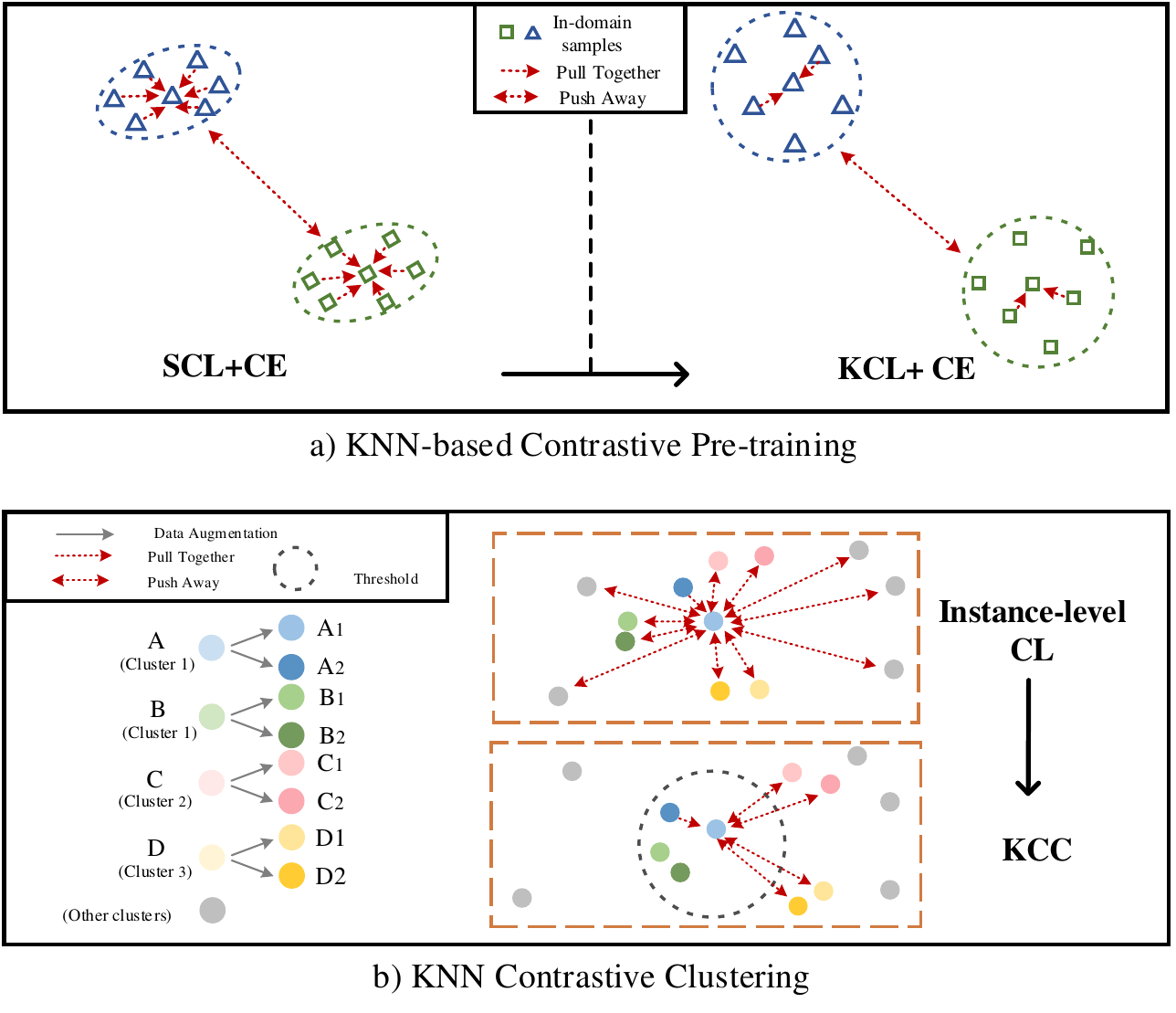}}
    \caption{The high-level idea of our KCOD compared to baselines. Fig (a) shows KCL increases the intra-class variance by only taking the K nearest samples of the same class as positive samples. Fig (b) displays that KCC filters out false negative samples that belong to the same class as the anchor and selects K nearest neighbors as true hard negatives to form clear cluster boundaries.}
    \label{intro}
\end{figure}

The key challenges of OOD intent discovery come from two aspects: (1) \textbf{Knowledge Transferability}. It requires transferring in-domain prior knowledge to help downstream OOD clustering. Early unsupervised intent discovery methods \cite{hakkani2015clustering, Padmasundari2018IntentDT, Shi2018AutoDialabelLD} only model unlabeled OOD data but ignore prior knowledge of labeled in-domain data thus suffer from poor performance. Then, recent work \cite{Lin2020DiscoveringNI, Zhang2021DiscoveringNI, mou-etal-2022-disentangled} focus more on the semi-supervised setting where they firstly pre-train an in-domain intent classifier then perform clustering algorithms on extracted OOD intent representations by the pre-trained IND intent classifier. For example, \citet{Lin2020DiscoveringNI, Zhang2021DiscoveringNI} pre-train a BERT-based \cite{devlin-etal-2019-bert} in-domain intent classifier using cross-entropy (CE) classification loss. \citet{mou-etal-2022-disentangled} further proposes a supervised contrastive learning (SCL) \cite{khosla2020supervised} objective to learn discriminative intent representations.
(2) \textbf{Jointly Learning Representations and Cluster Assignments}. It's important to learn OOD intent features while performing clustering. \citet{Lin2020DiscoveringNI} uses OOD representations to calculate the similarity of OOD sample pairs as weak supervised signals. The gap between pre-trained IND features and unseen OOD data makes it hard to generate high-quality pairwise pseudo labels. Then, \citet{Zhang2021DiscoveringNI} proposes an iterative clustering method, DeepAligned, to obtain pseudo cluster labels by K-means \cite{MacQueen1967SomeMF}. It performs representation learning and cluster assignment in a pipeline way. Further, \citet{mou-etal-2022-disentangled} introduces a multi-head contrastive clustering framework to jointly learn representations and cluster assignments using contrastive learning.

However, all of these methods still suffer from two problems: (1) \textbf{In-domain Overfitting}: The state-of-the-art (SOTA) OOD intent discovery methods \cite{Zhang2021DiscoveringNI, mou-etal-2022-disentangled} adopt general supervised pre-training objectives such as CE and SCL for IND pre-training, but none of them consider the following question: what kind of intent representation is more generalized to transfer to downstream OOD clustering? Although CE and SCL are effective for classifying known IND classes, such learned representations are poor for downstream transfer. \citet{zhao2020makes, feng2021rethinking} find larger intra-class diversity helps transfer. CE and SCL tend to pull all samples from the same class together to form a narrow intra-class distribution, thus ignore the intra-class diverse features, which makes the learned representations unfavorable to transfer to the downstream OOD clustering. (2) \textbf{Gap between Clustering Objectives and Representation Learning}. Learning OOD intent representations is key for achieving efficient clustering. \citet{Zhang2021DiscoveringNI} can't align learning intent features with clustering because the two processes are iterative in a pipeline way. \citet{mou-etal-2022-disentangled} further proposes DKT to jointly learn representation and cluster assignment using contrastive learning \cite{Gao2021SimCSESC, Yan2021ConSERTAC}. However, such a contrastive objective of learning intent features pushes apart representations of different samples in the same class, which impair the clustering goal where samples within the same class should be compact. 

To solve the two problems, we propose a unified \textbf{K}-Nearest Neighbor \textbf{C}ontrastive Learning framework for \textbf{O}OD \textbf{D}iscovery (\textbf{KCOD}). For the in-domain overfitting issue, we  
propose a simple \textbf{K}-nearest neighbor \textbf{C}ontrastive \textbf{L}earning objective (KCL) for IND pre-training in Fig \ref{intro} (a). Compared with SCL, we only take the K nearest samples of the same class as positive samples, which helps to increase the intra-class variance while maintaining a large inter-class variance. Larger intra-class diversity helps downstream transfer. For the gap between clustering and representation learning objectives, we propose a \textbf{K}-nearest neighbor \textbf{C}ontrastive \textbf{C}lustering method (KCC) for OOD clustering in Fig \ref{intro} (b). Traditional instance-wise contrastive learning only regards an anchor and its augmented sample as positive pairs and pushes apart representations of different samples even in the same class. In contrast, KCC firstly filters out false negative samples that belong to the same class as the anchor and then selects the K nearest neighbors as true negatives. We aim to mine high-confident hard negatives and learn compact intent representations for OOD clustering to form clear cluster boundaries.

Our contributions are three-fold: (1) We propose a unified K-nearest neighbor contrastive learning (KCOD) framework for OOD discovery, which aims to achieve better knowledge transfer and learn better clustering representations. (2) We propose a K-nearest neighbor contrastive learning (KCL) objective for IND pre-training, and a K-nearest neighbor contrastive clustering (KCC) method for OOD clustering, which solve "in-domain overfitting" problem and bridge the gap between clustering objectives and representation learning. (3) Experiments and analysis demonstrate the effectiveness of our method for OOD discovery.

\section{Approach}
\label{method}

\begin{figure*}[t]
    \centering
    \resizebox{1.0\linewidth}{!}{
    \includegraphics{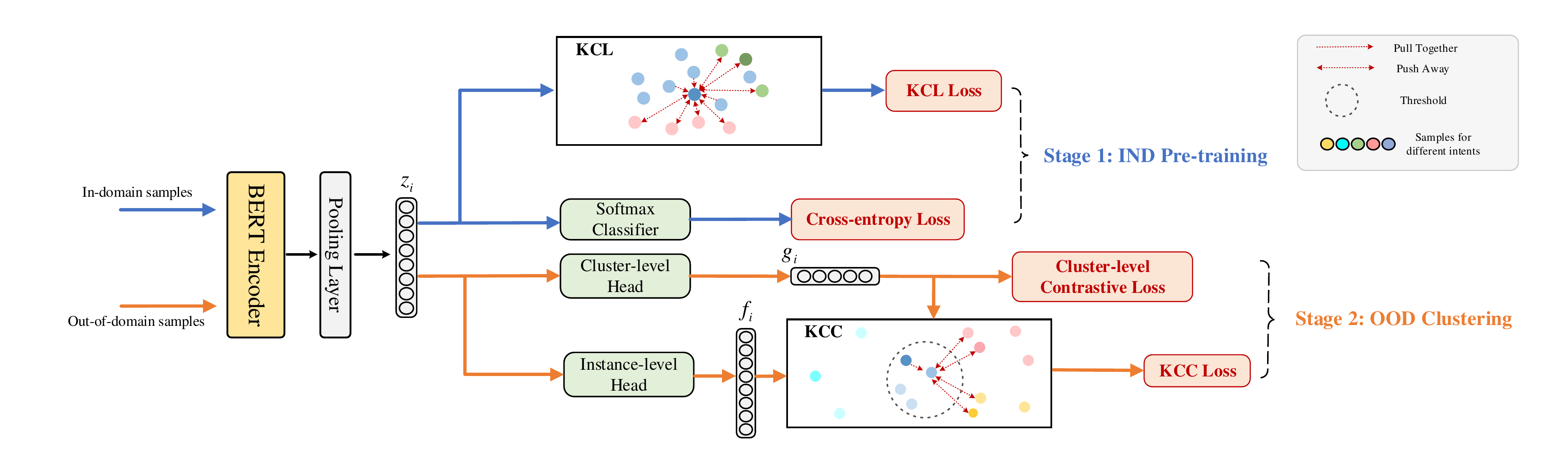}}
    \caption{The overall architecture of our proposed unified K-nearest neighbor contrastive learning framework for OOD discovery, KCOD. Stage 1 denotes IND pre-training and Stage 2 denotes OOD clustering.}
    \label{model}
\end{figure*}

OOD discovery assumes there is a set of labeled in-domain data  and unlabeled OOD data\footnote{We notice there are two settings of OOD discovery: one is to cluster unlabeled OOD data and another is to cluster unlabeled mixed IND\&OOD data. Here we adopt the first setting as \citet{mou-etal-2022-disentangled} because mixing IND\&OOD makes it hard to fairly evaluate the capability of discovering new intent concepts.}. Our goal is to cluster OOD concepts from unlabeled OOD data using prior knowledge from labeled IND data. The overall architecture of our KCOD is shown in Fig \ref{model}, including KNN contrastive IND pre-training and KNN contrastive OOD clustering. IND pre-training firstly gets generalized intent representations via our proposed KCL objective and then OOD clustering uses KCC to group OOD intents into different clusters.

\subsection{KNN Contrastive Pre-training}
K-nearest neighbor contrastive learning (KCL) aims to increase the intra-class variance to learn generalized intent representations for downstream clustering. Previous work \cite{zeng-etal-2021-modeling,mou-etal-2022-disentangled} pull together IND samples belonging to the same class and push apart samples from different classes. However, such methods make all the instances of the same class collapse into a narrow area near the class center, which reduces the intra-class diversity. \citet{zhao2020makes, feng2021rethinking} find large intra-class diversity helps transfer knowledge to downstream tasks. Therefore, we relax the constraint by only limiting k-nearest neighbors close to each other. The KCL pre-training loss is as follows:
{\setlength{\abovedisplayskip}{0.2cm}
\setlength{\belowdisplayskip}{0.2cm}
\begin{align}
\begin{split}
    \mathcal{L}_{K C L}=\sum_{i=1}^{N}-\frac{1}{|{K}_{i}|}
    \sum_{j=1}^{{K}_{i}} \log \frac{\exp \left(f_{i} \cdot f_{j} / \tau\right)}{\sum_{k=1}^{|A_{ij}|} \exp \left(f_{i}\cdot f_{k} / \tau\right)}
\end{split}
\end{align}}

where $K_{i}$ is the set of k-nearest neighbors with the same class as $i$-th sample. $A_{ij}$ denotes the union set of the positive $f_{j}$ and negative samples whose classes are different from the $i$-th sample. Specifically, given an anchor sample, we firstly get all the samples belonging to the same class as the anchor from the batch, then select its k-nearest neighbors among these samples using extracted intent features. KCL aims to pull together the anchor and its neighbors in the same class and push apart samples from different classes.
To support a large batch size, we employ a momentum queue \cite{He2020MomentumCF} to update the intent features. The queue decouples the size of contrastive samples from the batch size, resulting in a larger negative set. We perform joint training both using KCL and CE, and simply adding them gets the best performance. Section \ref{kcl} proves our proposed KCL increases the intra-class variance and alleviates in-domain overfitting.


\subsection{KNN Contrastive Clustering}
After obtaining pre-trained intent features, we need to group OOD intents into different clusters. Existing work \cite{Zhang2021DiscoveringNI,mou-etal-2022-disentangled} can't jointly learn representations and cluster assignments. \citet{Zhang2021DiscoveringNI} iteratively performs the two stages, leading to a suboptimal result. \citet{mou-etal-2022-disentangled} instead employs a contrastive clustering framework for joint learning but still has a gap between clustering objectives and representation learning. We first briefly introduce the contrastive clustering framework and then provide an analysis of how to bridge the gap.

Given an OOD example $x_{i}$, we firstly use the pre-trained BERT encoder to get an OOD intent feature $z_{i}$. Then, we use a cluster-level contrastive loss (CL) $\ell_{i, j}^{clu}$ to learn cluster assignments. Specifically, we project $z_{i}$ to a vector $g_{i}$ with dimension C which equals to the pre-defined cluster number\footnote{Estimating cluster number C is out of the scope of this paper. We provide a discussion in Section \ref{c}.}. So we get a feature matrix of $N \times C$ where N is the batch size. Following \citet{li2021contrastive}, we regard $i$-th column of the matrix as the $i$-th cluster representation $y_{i}$ and construct cluster-level loss $\ell_{i, j}^{clu}$ as follows:
{\setlength{\abovedisplayskip}{0.1cm}
\setlength{\belowdisplayskip}{0.2cm}
\begin{equation}
    \begin{aligned}
    \ell_{i, j}^{clu}=-\log \frac{\exp \left(\operatorname{sim}\left(y_{i}, y_{j}\right) / \tau\right)}{\sum_{k=1}^{2 C} 1_{[k \neq i]} \exp \left(\operatorname{sim}\left(y_{i}, y_{k}\right) / \tau\right)}
 \end{aligned}
\end{equation}
}
where $y_{j}$ is the dropout-augmented \cite{Gao2021SimCSESC} cluster representation of $y_{i}$ and $\operatorname{sim}$ denotes cosine distance. To learn intent representations, \citet{li2021contrastive,mou-etal-2022-disentangled} use an instance-level contrastive learning loss $\ell_{i, j}^{ins}$:
{\setlength{\abovedisplayskip}{0.2cm}
\setlength{\belowdisplayskip}{0.2cm}
\begin{equation}
    \begin{aligned}
     \ell_{i, j}^{ins}=-\log \frac{\exp \left(\operatorname{sim}\left(f_{i}, f_{j}\right) / \tau\right)}{\sum_{k=1}^{2 N} \mathbf{1}_{[k \neq i]} \exp \left(\operatorname{sim}\left(f_{i}, f_{k}\right) / \tau\right)} 
    \end{aligned}
\end{equation}
} 
where $f_{i}$ is transformed from $z_{i}$ by an instance-level head and $f_{j}$ denotes its dropout augmentation. $\tau$ is the temperature. However, Eq 3 only regards an anchor and its augmented sample as positive pair and even pushes apart different samples in the same class. The characteristic has a conflict with the clustering goal where the samples of the same class should be tight. This instance-level CL loss considers the relationship between instances instead of different types, which makes it hard to learn distinguished intent cluster representations. 

Therefore, we propose a K-nearest neighbor contrastive clustering method (KCC) to form clear cluster boundaries. Firstly, we use the predicted logits from the cluster-level head $g$ \footnote{The output dim of projector $g$ is equal to the cluster number C, so we can take the normalized output vector as the predicted probability on all clusters.} and compute the dot similarity of two samples to filter out false negative samples that belong to the same class as the anchor. Here, we find a simple similarity threshold can work well (see Section \ref{threshold}). So we select samples whose similarity scores are below the threshold as negatives. To further separate different clusters, we select the K nearest neighbors from the negative set as hard negatives. Our intuition is to push these hard negatives away from the anchor and form clear cluster boundaries. We formulate the KCC loss as follows:
{\setlength{\abovedisplayskip}{0.2cm}
\setlength{\belowdisplayskip}{0.2cm}
\begin{equation}
    \begin{aligned}
     \ell_{i, j}^{KCC}=-\log \frac{\exp \left(\operatorname{sim}\left(f_{i}, f_{j}\right) / \tau\right)}{  \sum_{k=1}^{|H_{i}|} \exp \left(\operatorname{sim}\left(f_{i}, f_{k}\right) / \tau\right) }
    \end{aligned}
\end{equation}
} where $H_{i}$ is the union set of the augmented positive sample $f_{j}$ and k-nearest hard negative set $N_{i}$ of $f_{i}$.
We give a theoretical explanation from the perspective of gradients. For convenience, we denote $s_{i, i}$ as the positive pair and $s_{i, j}, i \neq j$ as negative pairs. So original instance-level CL loss in Eq 3 is rewritten as:
{\setlength{\abovedisplayskip}{0.2cm}
\setlength{\belowdisplayskip}{0.2cm}
\begin{align}
    \mathcal{L}\left(x_{i}\right)=-\log \left[\frac{\exp \left(s_{i, i} / \tau\right)}{\sum_{k \neq i} \exp \left(s_{i, k} / \tau\right)+\exp \left(s_{i, i} / \tau\right)}\right]
\end{align}
}
where $s_{i, j}=sim(f\left(x_{i}\right), f\left(x_{j}\right))$. We analyze the gradients with respect to different negative samples following \citet{Wang2020UnderstandingCR}:
\begin{align}
\setlength{\abovedisplayskip}{0.1cm}
\setlength{\belowdisplayskip}{0.1cm}
    \quad \frac{\partial \mathcal{L}\left(x_{i}\right)}{\partial s_{i, j}} =\frac{1}{\tau} \frac{\exp \left(s_{i, j} / \tau\right)}{\sum_{k \neq i} \exp \left(s_{i, k} / \tau\right)+\exp \left(s_{i, i} / \tau\right)}
\end{align}
From Eq 5, we find that if easy negatives are filtered out, the gradient (penalty) to other hard negatives gets larger, thus pushing away these negatives from the anchor. It means our model can separate misleading samples near the cluster boundary (see Section \ref{kcc}). We simply add the cluster-level CL loss and KCC loss to jointly learn cluster assignments and intent representations. Following \citet{li2021contrastive}, we also add a regularization item to avoid the trivial solution that most instances are assigned to a single cluster. For inference, we only use the cluster-level head and compute the argmax to get the cluster results without additional K-means. 

\section{Experiments}
\label{main_results}

\subsection{Datasets}
We conduct experiments on three benchmark datasets, Banking \cite{casanueva-etal-2020-efficient}, HWU64 \cite{liu2021benchmarking} and CLINC \cite{larson-etal-2019-evaluation}. Banking contains 13,083 customer service queries with 77 intents in the banking domain. HWU64 includes 25,716 utterances with 64 intents across 21 domains. CLINC contains 22,500 queries covering 150 intents across 10 domains. Following \citet{mou-etal-2022-disentangled}, we randomly sample a ratio of the intents as OOD (10\%, 20\%, 30\% for Banking, 30\% for CLINC and HWU64), and the rest as IND. Note that we only use the IND data for pre-training and use OOD data for clustering. To avoid randomness, we average results over three random runs.



\begin{table*}[t]
\centering
\resizebox{1.0\textwidth}{!}{%
\begin{tabular}{ll|lll|lll|lll|lll|lll}
\hline
\multicolumn{2}{c|}{\multirow{2}{*}{Method}}                                          & \multicolumn{3}{c|}{Banking-10\%} & \multicolumn{3}{c|}{Banking-20\%} & \multicolumn{3}{c|}{Banking-30\%} &  \multicolumn{3}{c|}{HWU64-30\%}&  \multicolumn{3}{c}{CLINC-30\%}\\
                                                         &                & ACC      & ARI      & NMI      & ACC      & ARI      & NMI    & ACC  & ARI     & NMI  & ACC  & ARI     & NMI & ACC  & ARI     & NMI    \\ \hline
                                        
\multicolumn{2}{l|}{PTK-means(SCL)}               & 55.00	&36.18	&53.75    & 51.68    & 35.65     & 56.77    & 45.06    & 32.12     & 57.93     & 56.95	& 43.79	& 61.94   & 61.63    & 40.96    & 75.90\\
\multicolumn{2}{l|}{PTK-means(KCL)}              & 70.91 & 57.83 & 67.53 & 63.62  &53.69  &66.54 &59.60  &47.92  &66.78 &72.53  &58.56  &71.26 &83.17  &76.53  &87.80 \\
                                                  \multicolumn{2}{l|}{DeepCluster \cite{Caron2018DeepCF}}                     & 60.59	&41.88&	55.22    & 60.33    & 50.21    & 69.54    & 59.35     & 45.94    & 68.08    & 76.35	& 65.40	& 78.40 & 78.09    & 71.05    & 88.70\\
                                                  \multicolumn{2}{l|}{CDAC+ \cite{Lin2020DiscoveringNI}}                            &77.50	&60.53&	71.14    & 63.50     & 53.94     & 72.35    & 59.78    &44.58     &69.19     &75.08	&61.18	&79.51   & 73.04    & 64.44    & 87.90\\
                                                  \multicolumn{2}{l|}{DeepAligned \cite{Zhang2021DiscoveringNI}}                     & 77.78	&66.95	&76.91    &67.01     &58.79      &76.06     &63.86     &52.84     &73.66     &82.04	&76.13	&86.35    & 91.56    & 86.58    & 94.91\\
                                                  \multicolumn{2}{l|}{DKT \cite{mou-etal-2022-disentangled}}       &84.69    &71.11      &76.92     &69.55     &57.00     &73.21     &66.50     &52.07      &72.22 &83.91 	&73.69	&83.83   &94.96	&90.25	&95.94\\ \hline
                                                  \multicolumn{2}{l|}{KCOD w/o KCC(ours)}       &85.21    &71.67      &78.40     &71.07     &59.45     &74.71     &69.35     &54.78      &73.74 &84.55 	&76.54	&84.91   &95.62	&91.61	&96.67\\
                                                  
                                                  \multicolumn{2}{l|}{KCOD(ours)} & \textbf{86.67}    & \textbf{74.05}    & \textbf{79.89}    & \textbf{73.09}    & \textbf{60.96}    & 75.67    & \textbf{71.09}    & \textbf{57.73}     & \textbf{75.79} & \textbf{86.28}	& \textbf{77.07}	& 85.62 &\textbf{96.48}	&\textbf{92.46}	&\textbf{96.89}\\ \hline

\end{tabular}
}
\caption{Performance comparison on three datasets. For Banking, We randomly sample 10\%, 20\% and 30\% of all classes as OOD types. For HWU64 and CLINC, we randomly sample 30\% of all classes as OOD types. KCOD w/o KCC denotes we use the same clustering method as DKT and replace original SCL with KCL for IND pre-training. Results are averaged over three random runs. ($p < 0.01$ under t-test)}
\label{tab:main_result}
\end{table*}

\subsection{Baselines}
Similar with \citet{mou-etal-2022-disentangled}, we mainly compare our method with semi-supervised baselines: PTK-means \footnote{Here we conduct fair experiments with different IND pre-training objectives, in which PTK-means(SCL) adopts the same pre-training objective as DKT, and PTK-means(KCL) adopts the same pre-training objective as KCOD.} (k-means with IND pre-training), DeepCluster \cite{Caron2018DeepCF}, CDAC+ \cite{Lin2020DiscoveringNI}, DeepAligned \cite{Zhang2021DiscoveringNI} and DKT \cite{mou-etal-2022-disentangled}, in which DKT is the current state-of-the-art method for OOD intent discovery. We leave the details of the baselines in Appendix \ref{baselines}. For fairness, all baselines use the same BERT backbone. We adopt three widely used metrics to evaluate the clustering results: Accuracy (ACC), Normalized Mutual Information (NMI), and Adjusted Rand Index (ARI). ACC is the most important metric. Note that for Banking-10\% and CLINC-30\%, the results of all baselines except PTK-means(KCL) are retrieved from \citet{mou-etal-2022-disentangled}, while for Banking-20\%, Banking-30\% and HWU64-30\%, we rerun all baselines with the same dataset split for fair comparison. \footnote{\citet{mou-etal-2022-disentangled} focuses on the multi-domain dataset CLINC, however, we find CLINC is relatively simple for its coarse-grained intent types. In this paper, we focus on the more challenging single-domain fine-grained dataset Banking.}

\subsection{Main Results}
Table \ref{tab:main_result} shows the main results of our proposed method compared to the baselines. In general, our method consistently outperforms all the previous baselines with a large margin. We analyze the results from four aspects:

\noindent\textbf{Our proposed KCL objective helps knowledge transfer.} We can see that KCOD w/o KCC has a significant improvement compared to DKT and DeepAligned. For example, KCOD w/o KCC outperforms previous state-of-the-art DKT by 2.85\% (ACC), 2.71\%(ARI), 1.52\%(NMI) on Banking-30\%. It is worth noting that KCOD w/o KCC adopts the same OOD clustering method as DKT, but uses our proposed KCL objective instead of SCL for IND pre-training. It proves that using the KCL pre-training objective learns generalized intent representations, which helps transfer in-domain prior knowledge for downstream OOD clustering. We also provide a deep analysis in Section \ref{kcl} to explore the reasons.

\noindent\textbf{Our proposed KCC helps OOD clustering.} We can observe that KCOD further improves 1.74\%(ACC), 2.95\%(ARI) and 2.05\%(NMI) compared to KCOD w/o KCC on Banking-30\%. This proves that KCC can learn cluster-friendly representation, which is helpful for OOD clustering. We discuss the effect of KCC in detail in Section \ref{kcc}.


\begin{table}[t]
\centering
\small
\resizebox{0.42\textwidth}{!}{
\begin{tabular}{l|c|c|c}
\hline
               & ACC   & ARI   & NMI    \\ \hline
No-pretraining & 32.99 & 16.59 & 33.47 \\
CE            & 67.65 & 52.75 & 70.37 \\ 
CE+SCL             & 69.55 & 57.00 & 73.21  \\ \hline
CE+KCL         & \textbf{71.07} & \textbf{59.45} & \textbf{74.71} \\ \hline
\end{tabular}}

\caption{Clustering performance comparison of different pre-training objectives using the same clustering method. We use Banking-20\% for analysis.}
\label{tab:kcl_cluster}
\end{table}

\begin{figure*}[t]
    \centering
    \subfigure[SCL\_IND]{
        \includegraphics[scale=0.280]{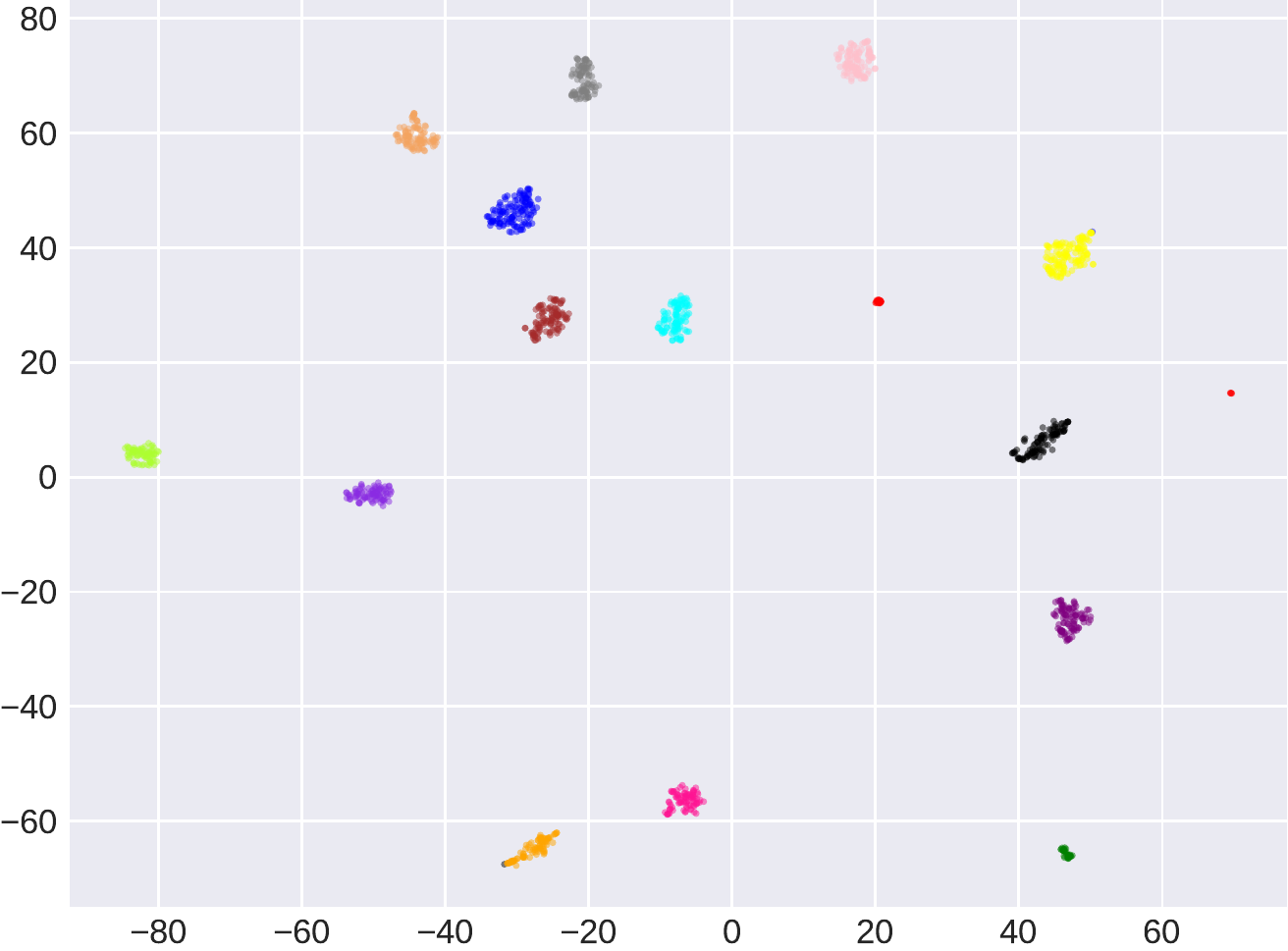}
    }
    \subfigure[KCL\_IND]{
        \includegraphics[scale=0.280]{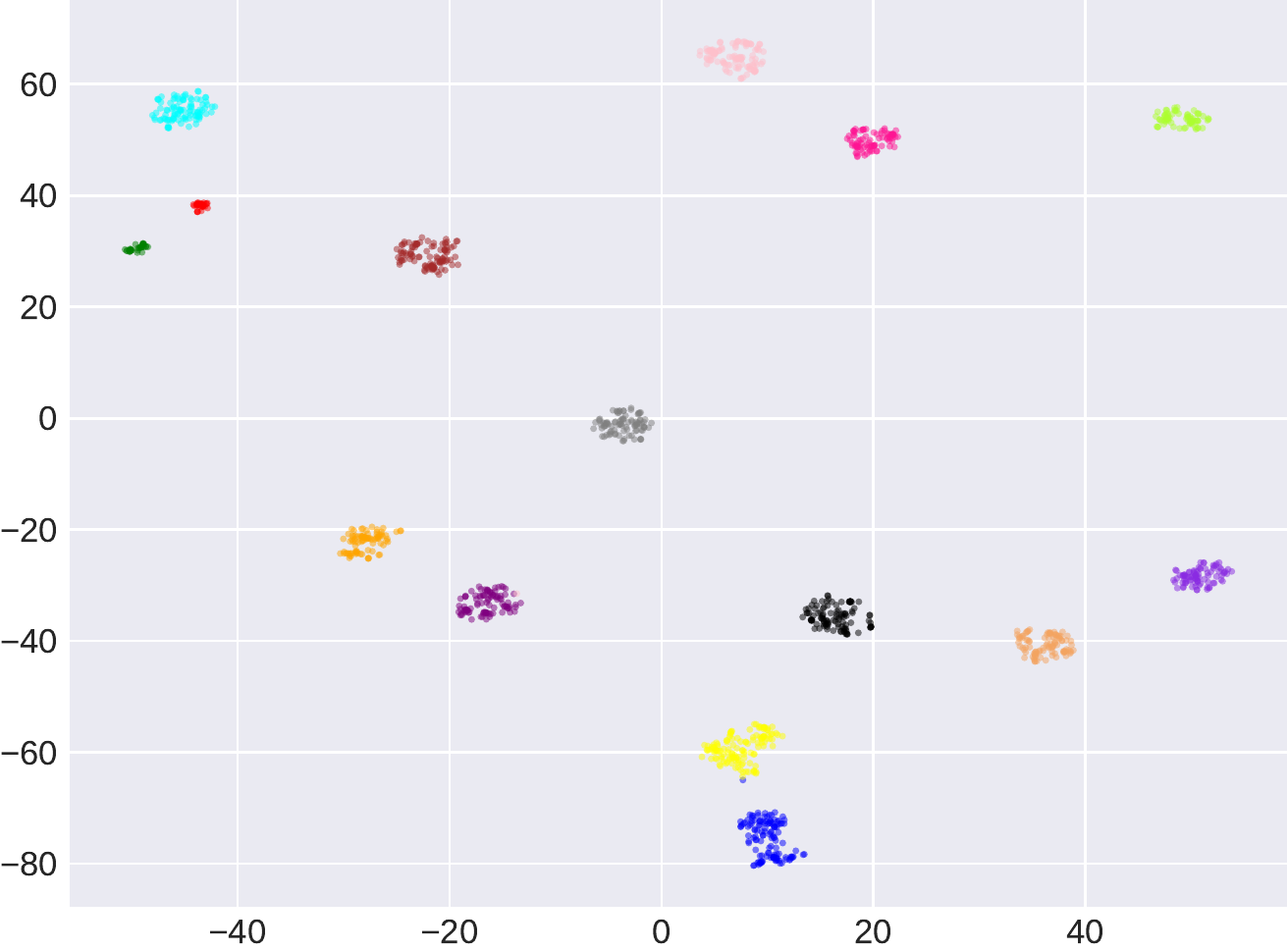}
    }
    \subfigure[SCL\_OOD]{
        \includegraphics[scale=0.280]{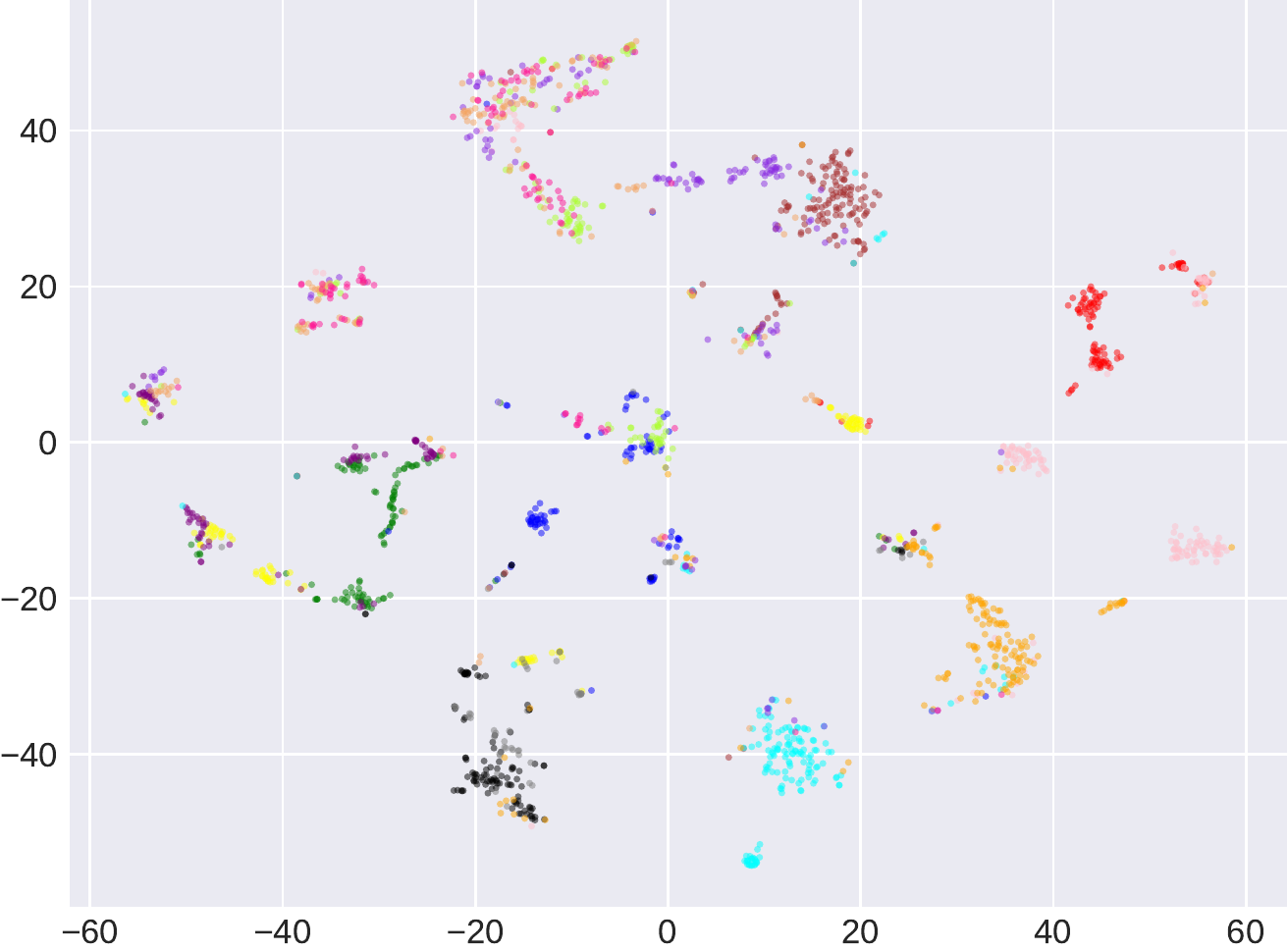}
    }
    \subfigure[KCL\_OOD]{
        \includegraphics[scale=0.280]{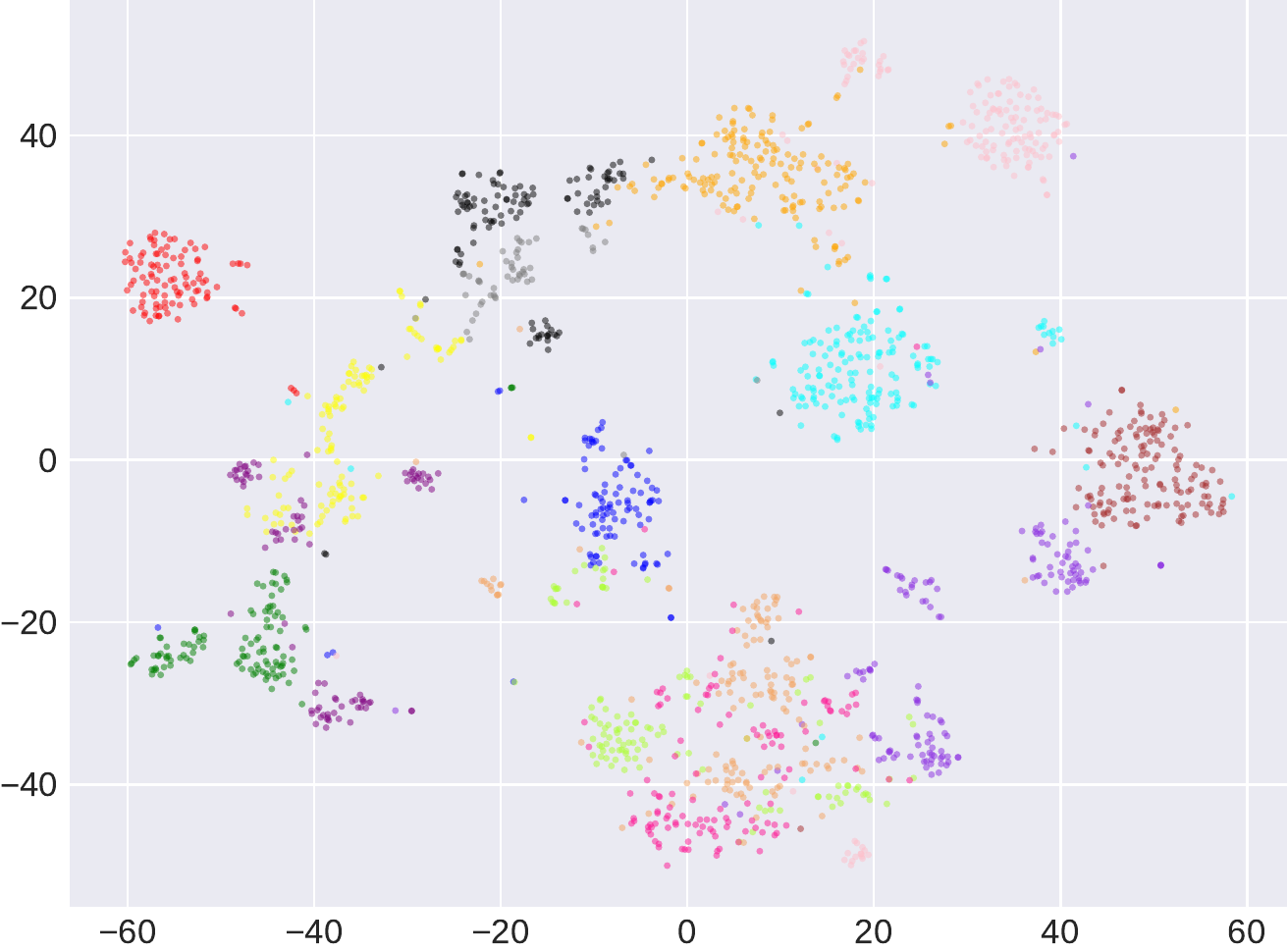}
    }
    \caption{IND and OOD intent visualization for different IND pre-training objectives.}
    \label{fig:Visualization_pretraining}
\end{figure*}

\noindent\textbf{Comparison of different datasets} We validate the effectiveness of our method on different datasets, where Banking is a single-domain fine-grained dataset, and CLINC and HWU64 are multi-domain datasets. We can see that all methods perform significantly worse on Banking than on CLINC and HWU64 datasets, which indicates that the single-domain fine-grained scenario is more challenging for OOD discovery. But our proposed KCOD achieves larger improvements of 3\%-6\% on the Banking-30\% dataset compared to DKT, while only 2\%-4\% on CLINC-30\% and HWU64-30\%. It indicates that KCOD can better cope with the challenges in fine-grained intent scenarios, and has stronger transferability and generalization.

\noindent\textbf{Effect of different OOD ratios} We observe that all methods decrease significantly when the OOD ratio increases. Because the proportion of unlabeled OOD increases and labeled IND data decreases, making both knowledge transfer and clustering more difficult. However, our KCOD achieves more significant improvements with the increase of the OOD ratio. For example, compared to DKT, on Banking-10\%, KCOD increases by 1.98\% (ACC), on Banking-20\%, KCOD increases by 3.54\% (ACC), and on Banking-30\%, KCOD increases by 4.59\% (ACC). This also reflects the strong generalization capability of the KCOD framework.

\section{Qualitative Analysis}
\label{analysis}

\begin{figure*}[t]
    \centering
    \centering
    \subfigure[DKT]{
        \includegraphics[scale=0.375]{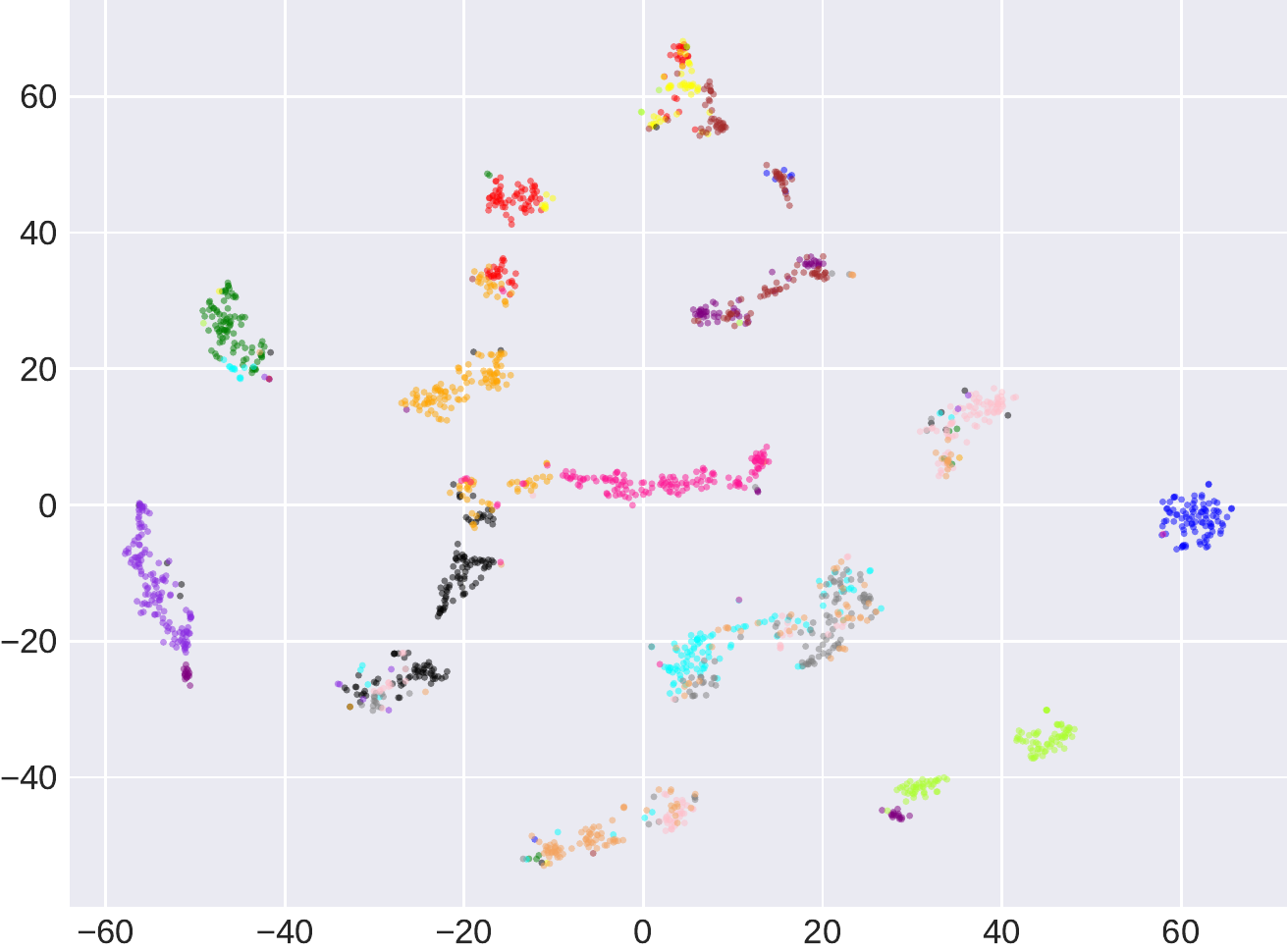}
    }
    \subfigure[KCOD w/o KCC(ours)]{
        \includegraphics[scale=0.375]{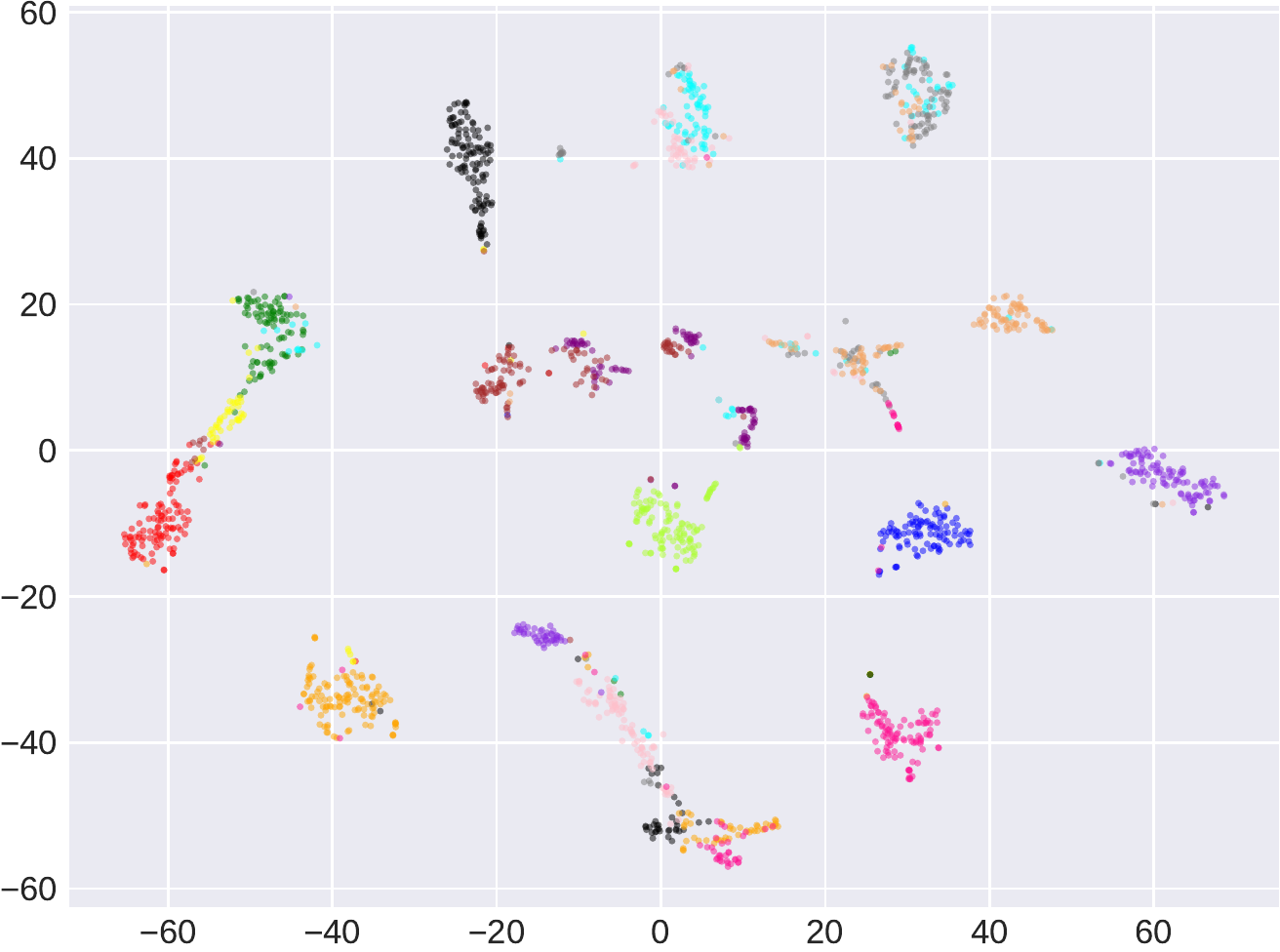}
    }
    \subfigure[KCOD(ours)]{
        \includegraphics[scale=0.375]{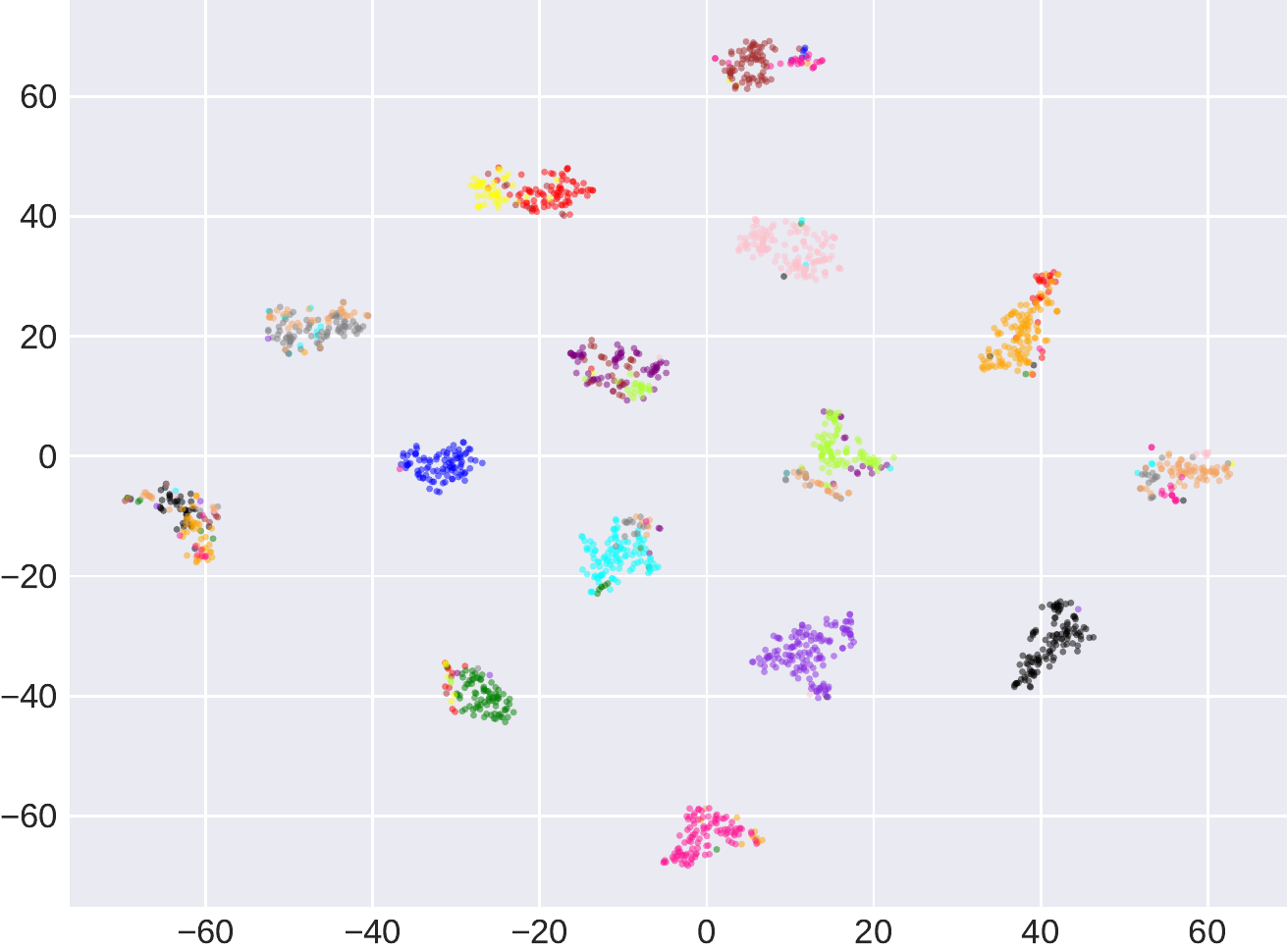}
    }
    \caption{OOD intent visualization of different models. We use the same OOD test set of Banking-20\%.}
    \label{Effect of KCC}
\end{figure*}

\subsection{Effect of KCL}
\label{kcl}

We analyze the effect of KCL from multiple perspectives.

\noindent\textbf{Ablation Study} We compare the OOD clustering performance under different IND pre-training objectives in Table \ref{tab:kcl_cluster}. All models employ the same contrastive clustering method as DKT \cite{mou-etal-2022-disentangled} for OOD clustering. We find that adding the KCL objective for IND pre-training significantly improves the performance for OOD clustering compared to SCL, which proves that KCL enhances knowledge transferability.

\noindent\textbf{Discussion of why KCL is effective} To explore why KCL is effective for knowledge transferability, we calculate the intra-class and inter-class distances following \citet{feng2021rethinking}. For the intra-class distance, we calculate the cosine similarity between each sample and its class center. For the inter-class distance, we calculate the cosine similarity between each class center and its 3 nearest class centers. We report the the averaged $1-\cos(\cdot ,\cdot )$ in Table \ref{tab:distance}. Results show that using KCL for IND pre-training can increase the intra-class distance of IND classes while maintaining a relatively large inter-class variance. Then we extract the representation of the OOD intents using the pre-trained model and perform K-means \cite{MacQueen1967SomeMF} (see K-means ACC in Table \ref{tab:distance}). We find KCL for IND pre-training helps OOD clustering and benefits knowledge transfer.

\noindent\textbf{Visualization} Fig \ref{fig:Visualization_pretraining} displays IND and OOD intent visualization for different IND pre-training methods SCL and KCL. We find KCL is beneficial to increasing the intra-class variance and helps build the clear OOD boundary. We argue that too small intra-class variance or too small inter-class variance are not good for downstream transfer, and preserving intra-class diverse features is vital to generalization.

In summary, our proposed KCL increases the intra-class variance and preserves the features related to intra-class difference by selecting the top K positives, which alleviates the "in-domain overfitting" problem and helps knowledge generalization to OOD clustering.

\begin{table}[t]
\centering
\resizebox{0.46\textwidth}{!}{%
\begin{tabular}{l|cc|c}
\hline

\multicolumn{1}{c|}{} & Intra-class $\mathbf{\uparrow}$ & Inter-class $\mathbf{\uparrow}$ & K-means ACC $\mathbf{\uparrow}$   \\ \hline
CE                   & 0.04        & 0.24        &  58.11    \\ 
CE+SCL                    & 0.01        & \textbf{0.68}        & 51.68 \\ \hline
CE+KCL                & \textbf{0.10}        & 0.43        & \textbf{63.62} \\ \hline
\end{tabular}%
}
\caption{Representation distribution of different pre-training objectives.}
\label{tab:distance}
\end{table}

\begin{table}[t]
\centering
\resizebox{0.45\textwidth}{!}{%
\begin{tabular}{l|c|c|c}
\hline
      Models         & ACC   & ARI   & NMI   \\ \hline
KCC    & \textbf{73.09}   & \textbf{60.96}   & \textbf{75.67} \\
\quad-w/o instance-level head       & 69.50   & 55.27   & 69.96 \\
\quad-w/o cluster-level head       & 63.50   & 43.42   & 66.63 \\
 \hline
\end{tabular}%
}
\caption{Ablation study of branches for KCC (We use KCL objective for IND pre-training).}
\label{tab:contrastive_obj}
\end{table}

\subsection{Effect of KCC}
\label{kcc}

To study the effect of KCC, we perform OOD visualization of DKT, KCOD w/o KCC, and KCOD in Fig \ref{Effect of KCC}. We see that KCOD can form clear cluster boundaries and separate different OOD clusters.

\begin{figure}[t]
    \centering
    \resizebox{.40\textwidth}{!}{
    \includegraphics{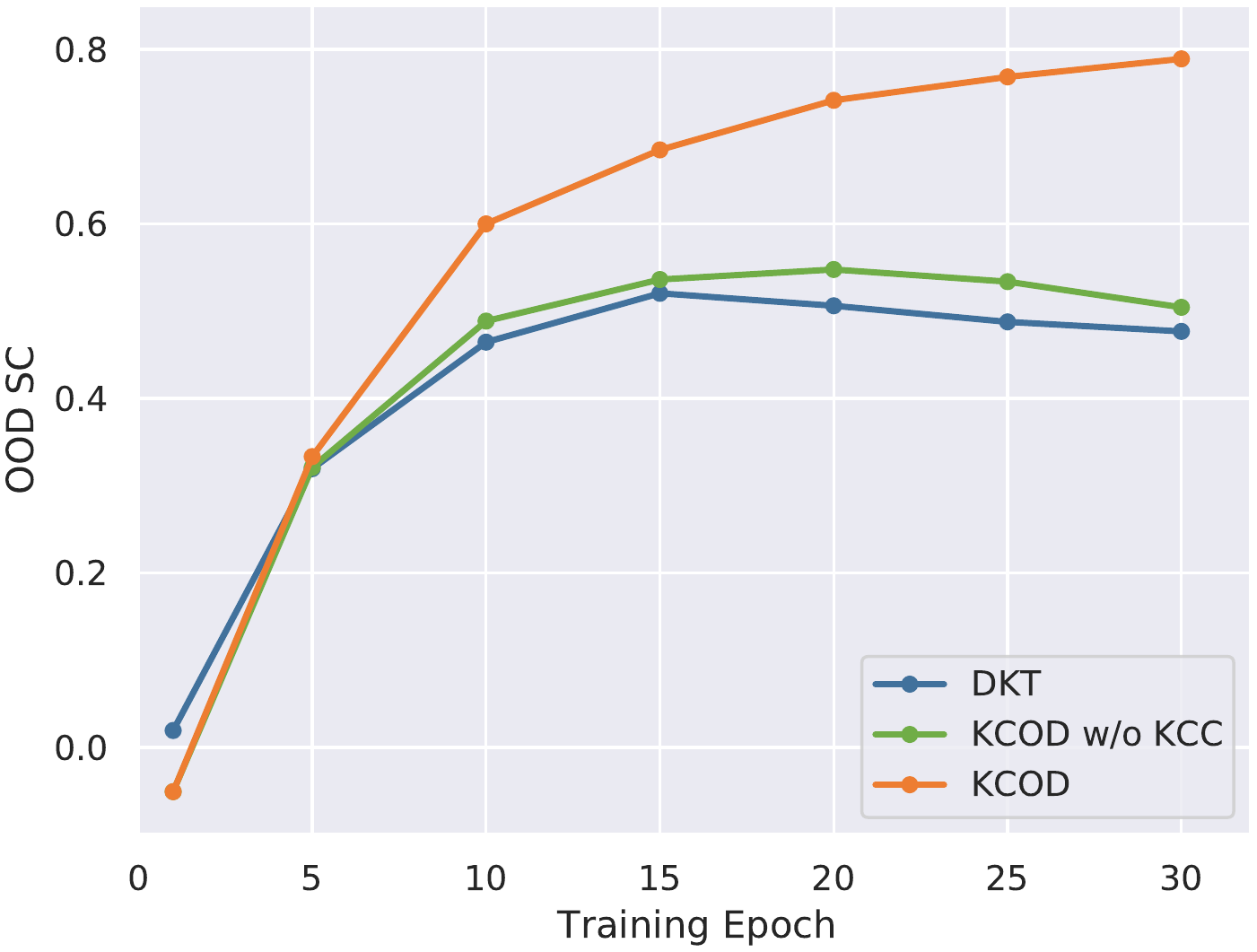}}
    \caption{OOD SC curves in the training process.}
    \label{SC_curve}
\end{figure}

\noindent\textbf{Bridge the gap between clustering and representation learning} We also show the curve of the OOD SC value during the training process in Fig \ref{SC_curve}. The range of SC is between -1 and 1, and the higher score means the better clustering quality \footnote{Please refer to more details about SC in Appendix \ref{sc}.}. Results show that previous contrastive clustering methods \cite{li2021contrastive, mou-etal-2022-disentangled} for OOD clustering, such as DKT and KCOD w/o KCC make the SC value first rise to a peak and then decrease to a certain extent. It's because they use instance-level contrastive learning(CL) to learn intent features which violates the clustering objective. Instance-level CL pulls together an anchor and its augmented positives and pushes apart representations of different samples, which means OOD intents within the same class are still separated. But clustering requires compact clusters. Our proposed KCC bridges the gap between clustering objectives and representation learning.

\begin{figure}[t]
    \centering
    \resizebox{.40\textwidth}{!}{
   \includegraphics{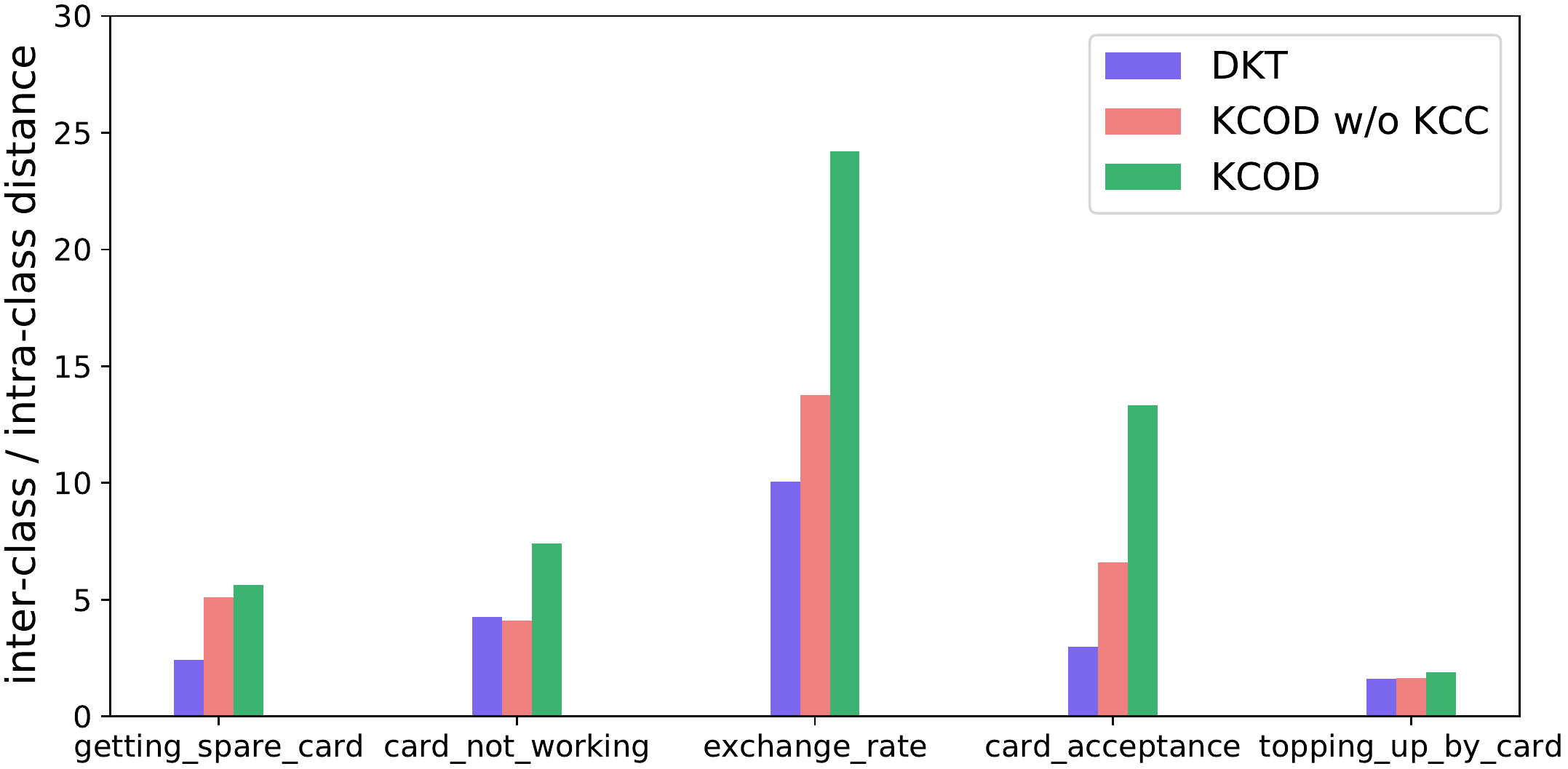}}
    \caption{Cluster compactness of different OOD discovery methods for 5 OOD classes.}
    \label{hard_class}
\end{figure}

\noindent\textbf{Form compact clusters} In order to more directly prove that KCC can effectively mine hard negative samples to form clear cluster boundaries and compact clusters, we randomly select five OOD classes in Banking-20\%, and calculate the ratio of inter-class distance and intra-class distance for different clustering methods respectively, as shown in Fig \ref{hard_class}. It can be seen that KCOD can form a more compact cluster for each class, which is beneficial to distinguishing these categories.

\subsection{Ablation Study for KCC}

K-nearest neighbor contrastive clustering (KCC) includes two branches, cluster-level head and instance-level head. In order to verify the effectiveness of the two branches working together for clustering. We performed an ablation study and the results are shown in Table \ref{tab:contrastive_obj}. For KCC w/o instance-level head, we remove the instance-level head and only use the cluster-level head for clustering. For KCC w/o cluster-level head, we remove the cluster-level head, and in the OOD clustering stage, instance-level contrastive learning is used for representation learning, and K-means is used for clustering.
It can be seen that the clustering performance drops significantly when any head is removed, which indicates that jointly learning instance-level representation and cluster-level assignments is beneficial for improving clustering performance.


\subsection{Hyper-parameter Analysis}
\label{hyper}

\begin{figure}[t]
    \centering
    \resizebox{.40\textwidth}{!}{
    \includegraphics{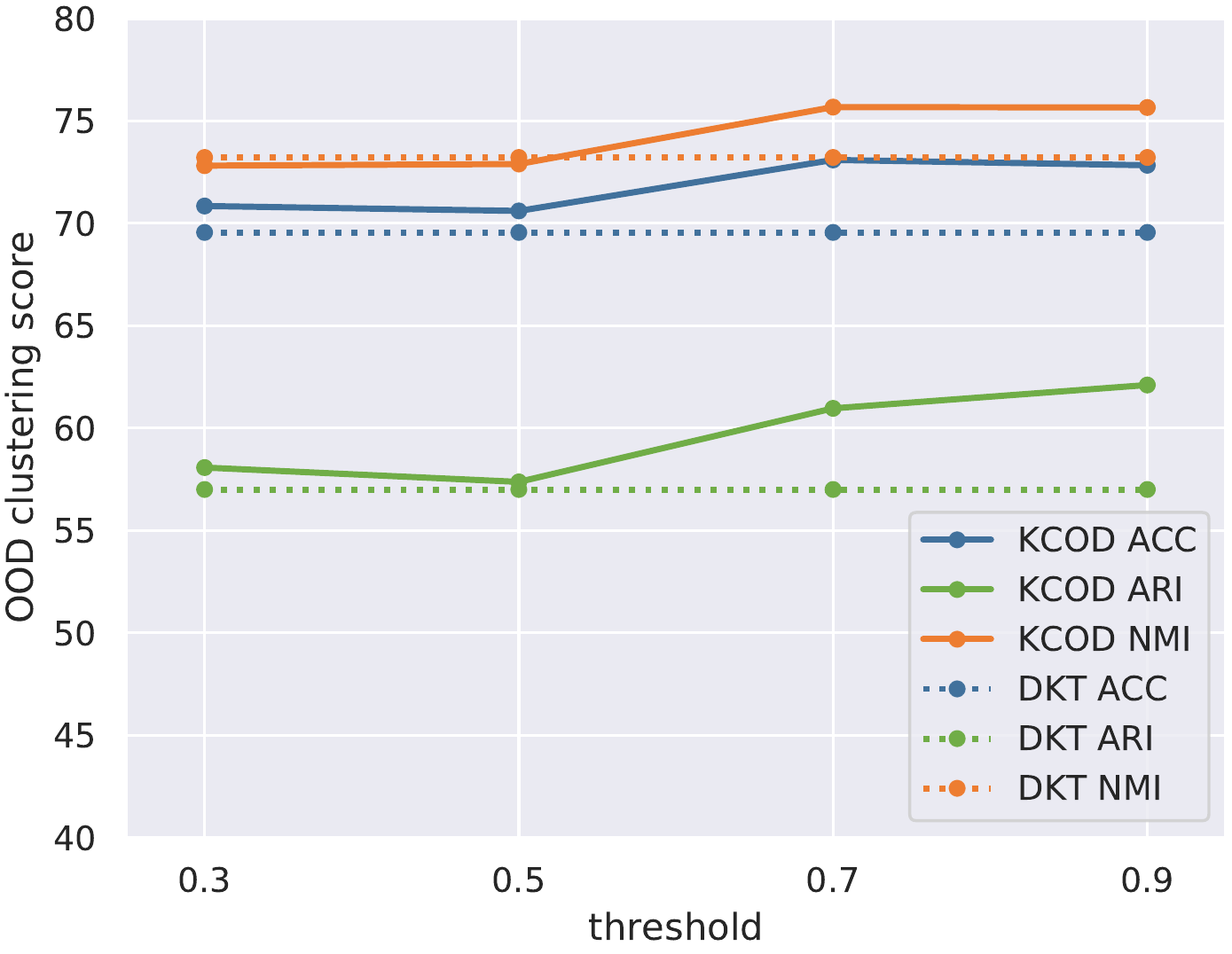}}
    \caption{Effect of the threshold of KCC on Banking-20\%. The range of $t$ is between 0 and 1.}
    \label{KCC_t}
\end{figure}

\begin{table}[t]
\centering
\small
\resizebox{0.38\textwidth}{!}{%
\begin{tabular}{l|c|c|c}
\hline
               & ACC   & ARI   & NMI    \\ \hline
DKT & 69.55 & 57.00 & 73.21 \\
+KCL (K=1)  &71.00	&56.91	&73.36 \\ 
+KCL (K=3)            & \textbf{71.07} & \textbf{59.45} & \textbf{74.71} \\ 
+KCL (K=5)               & 70.37 & 56.99 & 72.22  \\ 
+KCL (K=7)  & 69.33 & 56.69   & 73.03  \\ 
+KCL (K=9)           & 68.17 & 57.05 & 73.37 \\ \hline
\end{tabular}%
}
\caption{The effect of different K values of KCL.}
\label{tab:kcl_K}
\vspace{-0.3cm}
\end{table}


\noindent\textbf{The effect of the K value for KCL} Table \ref{tab:kcl_K} shows the effect of different K values of KCL pre-training loss. For a fair comparison, we replace SCL of DKT with our proposed KCL and use the same clustering method as DKT. We find smaller K achieves superior performance for OOD discovery because smaller K makes the model learn more intra-class diversity. But when K=1 KCL will be degraded to traditional instance-level contrastive learning and    lose label information.


\label{threshold}
\noindent\textbf{The effect of KCC threshold} Our proposed KCC method needs to firstly filter out false negative samples whose similarity with the anchor is greater than the specified threshold $t$. These false negatives are considered as candidate positive samples and should not be pushed apart like conventional instance-level contrastive learning \cite{Yan2021ConSERTAC, Gao2021SimCSESC, mou-etal-2022-disentangled}. We show the effect of different thresholds $t$ on the performance of KCOD in Fig \ref{KCC_t}. We find that when the threshold $t$ is in the range of [0.6, 0.9], the performance of KCOD is better than DKT. We argue that the KCC method needs to select a large threshold because a smaller threshold means that there may be more noise in the candidate positive sample set where some hard negative samples may be regarded as false positive samples.

\begin{figure}[t]
    \centering
    \resizebox{.40\textwidth}{!}{
    \includegraphics{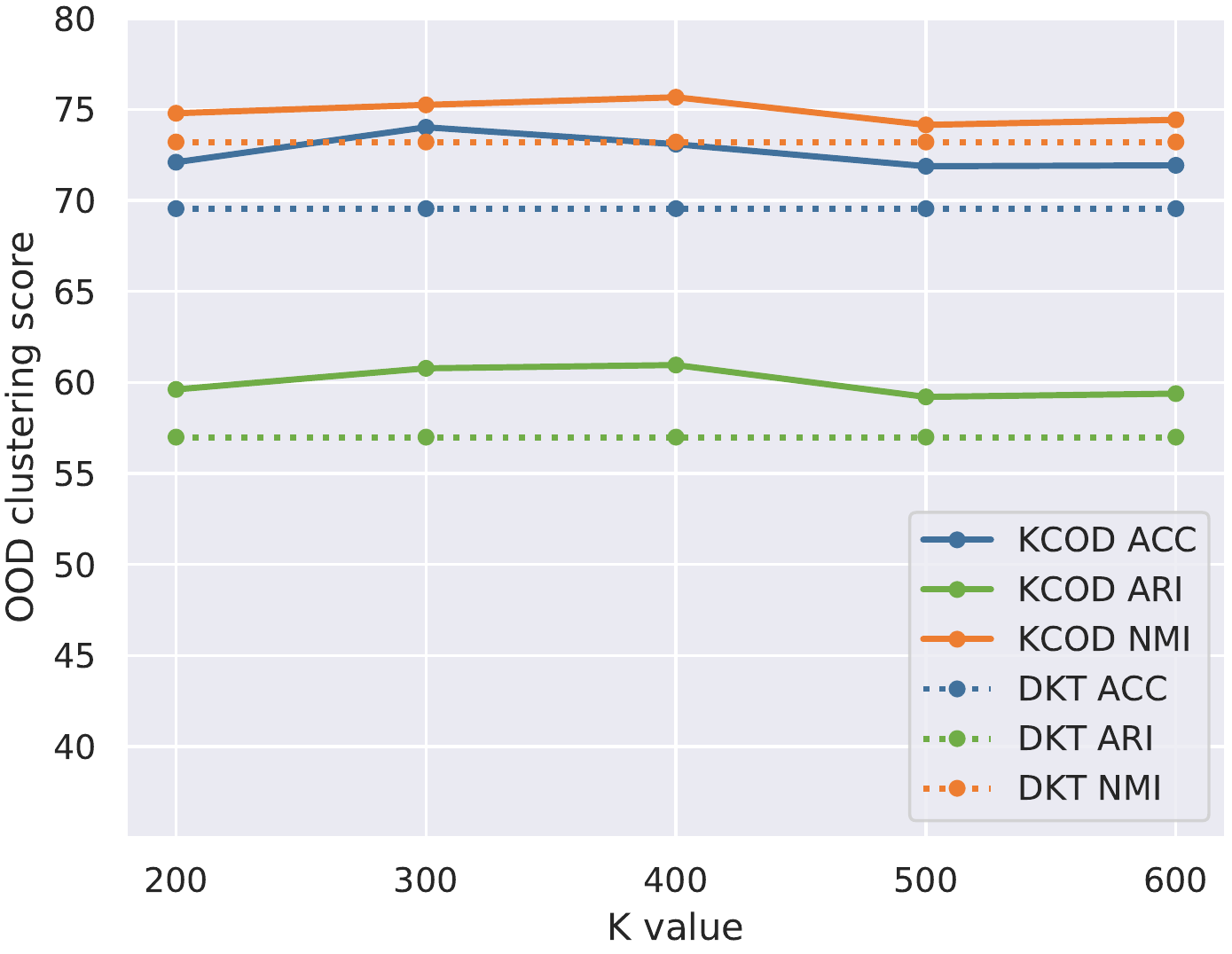}}
    \caption{Effect of the K value of KCC on Banking-20\%.}
    \label{KCC_K}
\end{figure}

\noindent\textbf{The effect of the K value of KCC} A core mechanism of KCC is to select K nearest neighbor samples in the candidate negative sample set to participate in the calculation of the contrastive learning loss. The main motivation is to mine hard negative samples to form clear cluster boundaries. We show the effect of different K values on KCOD performance in Fig \ref{KCC_K}. We find that our KCOD achieves consistent improvements under different K values, which proves the robustness of our method. And $K \in [300, 400]$  gets the best metrics. Too large K brings a subtle drop because many easy negatives are considered and hard negatives near the cluster boundary can't be explicitly separated. 

\subsection{Estimate the Number of Cluster C}
\label{c}

All the results we showed so far assume that the number of OOD classes is pre-defined. However, in real-world applications, the number of clusters often needs to be estimated automatically. Table \ref{tab:k} shows the results using the same cluster number estimation strategy \footnote{Here we use the same estimation algorithm as \citet{Zhang2021DiscoveringNI}. We leave the details in Appendix \ref{c}.}.  It can be seen that when the number of clusters is inaccurate, all methods have a certain decline, but our KCOD method still significantly outperforms all baselines, which also proves that KCOD is robust.

\begin{table}[]
\centering
\resizebox{0.38\textwidth}{!}{%
\begin{tabular}{l|c|c|c|c}
\hline
\multicolumn{1}{c|}{} & ACC     & ARI     & NMI    & C  \\ \hline
DeepAligned             & 67.00   & 58.79  & \textbf{76.06}  & 15 \\
DKT       & 69.55   & 57.00  & 73.21  & 15 \\
KCOD(ours)       & \textbf{73.09}   & \textbf{60.96}    & 75.67   & 15   \\ \hline
DeepAligned           & 66.50   & 58.00   & \textbf{75.00} & 13 \\
DKT     & 66.89   & 52.81   & 70.05   & 13 \\
KCOD(ours)      & \textbf{70.55}   & \textbf{58.62}   & 74.12  & 13   \\ \hline
\end{tabular}
}
\caption{Estimate the number of OOD clusters. C=13 is the estimated number compared to golden 15.}
\label{tab:k}
\end{table}

\subsection{Error Analysis}

\begin{figure*}[t]
    \centering
    \resizebox{0.8\textwidth}{!}{
    \includegraphics{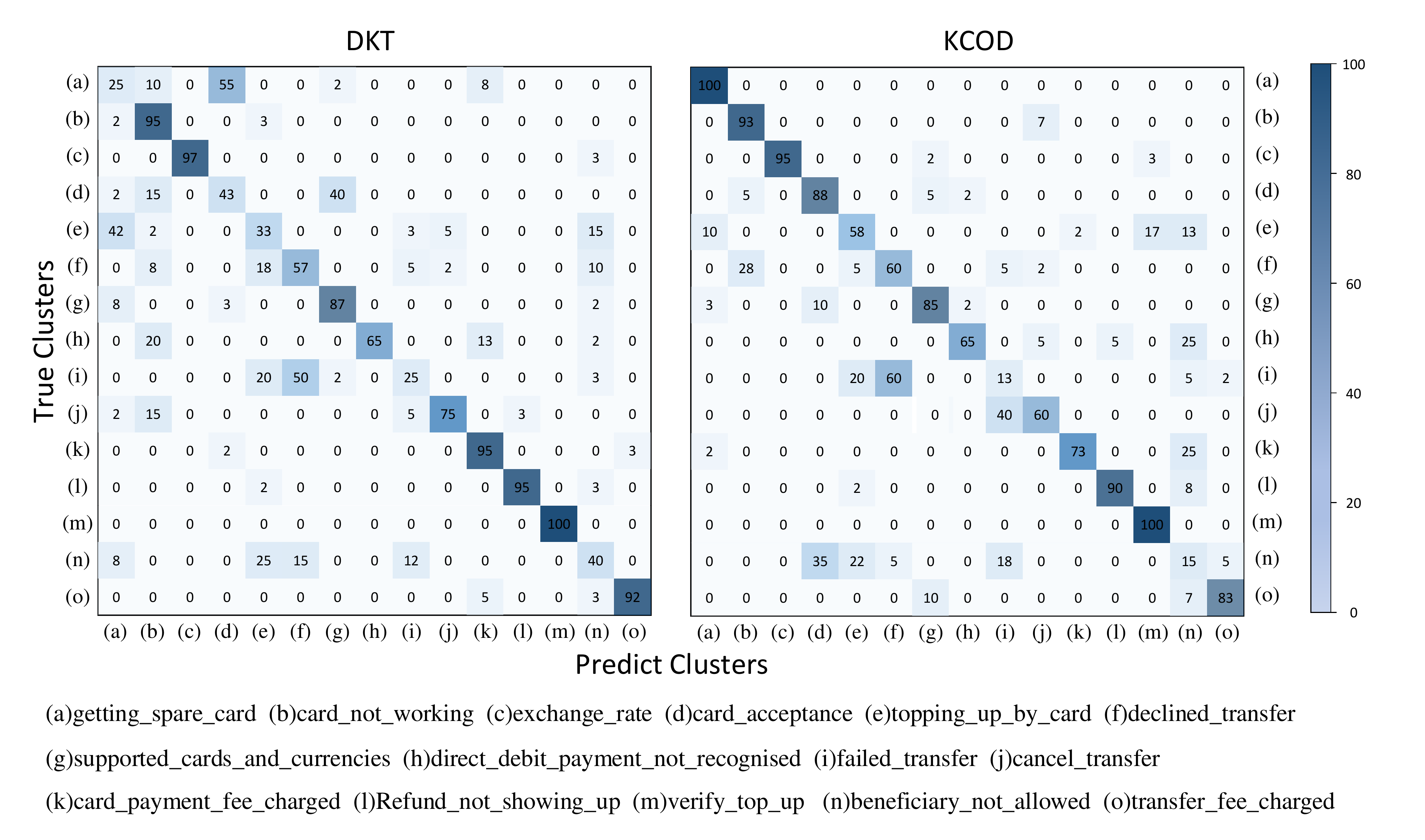}}
    \caption{Confusion matrix for the clustering results of  DKT and KCOD on Banking-20\%. The percentage values along the diagonal represent how many samples are correctly clustered into the corresponding class. The larger the number, the deeper the color.}
    \label{matrix}
\end{figure*}

We further analyze the error cases of DKT and KCOD in Fig \ref{matrix}. We find that for similar OOD intents, DKT is probably confused but our KCOD can effectively distinguish them. For example, DKT incorrectly groups \emph{getting\_spare\_card} intents into \emph{card\_acceptance} (55\% error rate) vs KCOD(0\%), which proves KCOD helps separate semantically similar OOD intents.

\section{Related Work}

\noindent\textbf{OOD Discovery} Early methods \cite{Xie2016UnsupervisedDE,Caron2018DeepCF} use unsupervised data for clustering. Recent work \cite{Lin2020DiscoveringNI, Zhang2021DiscoveringNI, mou-etal-2022-disentangled} performs semi-supervised clustering using labeled in-domain data. To transfer intent representation, \citet{Lin2020DiscoveringNI, Zhang2021DiscoveringNI} pre-train a BERT encoder using cross-entropy loss, then \citet{mou-etal-2022-disentangled} uses SCL \cite{khosla2020supervised} to learn discriminative features. All the models face the challenge of in-domain overfitting issues where representations learned from IND data will degrade for OOD data. Thus, we propose a KCL loss to keep large inter-class variance and help downstream transfer. For OOD clustering, \citet{Zhang2021DiscoveringNI,Zhang2022NewID} use k-means to learn cluster assignments but ignore joint learning intent representations. \citet{mou-etal-2022-disentangled} uses contrastive clustering where the instance-level contrastive loss for learning intent features has a gap with the cluster-level loss for clustering. Therefore, we propose the KCC method to mine hard negatives to form clear cluster boundaries. 

\noindent\textbf{Contrastive Learning} Contrastive learning (CL) is widely used in self-supervised learning \cite{He2020MomentumCF,Gao2021SimCSESC,khosla2020supervised}. \citet{zeng-etal-2021-modeling,Liu2021SelftrainingWM,Zeng2021AdversarialSL} apply it to OOD detection. \citet{Zhou2022KNNContrastiveLF} proposes a KNN-contrastive learning method for OOD detection. It aims to learn discriminative semantic features that are more conducive to anomaly detection. In contrast, our method uses a unified K-nearest neighbor contrastive Learning framework for OOD discovery where KCL increases intra-class diversity and helps downstream transfer, and KCC learns compact intent representations for OOD clustering to form clear cluster boundaries. \citet{mou-etal-2022-disentangled} uses contrastive clustering for OOD discovery. But original instance-level CL pushes apart different instances of the same intent which is against clustering. Thus, we use a simple k-nearest sampling mechanism to separate clusters and form clear boundaries.

\section{Conclusion}
In this paper, we propose a unified K-nearest neighbor contrastive learning (KCOD) framework for OOD intent discovery. We design a KCL objective for IND pre-training, and a KCC method for OOD clustering. Experiments on three benchmark datasets prove the effectiveness of our method. And extensive analyses demonstrate that KCL is helpful for learning intra-class diversity knowledge and alleviating the problem of intra-domain overfitting, and KCC is beneficial for forming compact clusters, effectively bridging the gap between clustering and representation learning. We hope to explore more self-supervised learning methods for OOD discovery in the future.


\section*{Limitation}
This paper mainly focuses on the out-of-domain (OOD) intent discovery task in task-oriented dialogue systems. We aims to leverage the prior knowledge of known in-domain (IND) intents to help OOD clustering.
Our proposed KCOD method well addresses the two challenges of knowledge transferability and joint learning of representation and cluster assignment, and achieves SOTA performance on three intent recognition benchmark datasets. However, our method can also be used in broader fields, such as short text clustering, topic discovery, etc., which we did not explore further in this paper. We will try to apply this framework to a wider range of NLP topics in the future.

\section*{Acknowledgements}
We thank all anonymous reviewers for their helpful comments and suggestions. This work was partially supported by National Key R\&D Program of China No. 2019YFF0303300 and Subject II No. 2019YFF0303302, DOCOMO Beijing Communications Laboratories Co., Ltd, MoE-CMCC "Artifical Intelligence" Project No. MCM20190701.

\bibliography{anthology,custom}
\bibliographystyle{acl_natbib}

\appendix


\section{Datasets}

\begin{table*}[t]
		\centering
		\resizebox{.9\textwidth}{!}{
		\begin{tabular}{ ccccccc }
			\toprule
			Dataset & Classes & Training & Validation & Test & Vocabulary & Length (max / mean) \\
			\midrule
			BANKING & 77  & 9,003 & 1,000 & 3,080 & 5,028 & 79 / 11.91 \\ 
			CLINC & 150  & 18,000 & 2,250 & 2,250 & 7,283 & 28 / 8.31 \\
			HWU64 & 64 & 8954 & 1076 & 1076  &4,948  &  25 / 6.57 \\
			\bottomrule
		\end{tabular}}
		\caption{Statistics of BANKING, CLINC and HWU64 datasets.}
		\label{tab:dataset1}
\end{table*}

We show the detailed statistics of Banking, HWU64 and CLINC datasets in Table \ref{tab:dataset1}.

\section{Baselines}
\label{baselines}
The details of baselines are as follows: 
\begin{itemize}
    \item \textbf{PTK-means} This method pre-trains the encoder network with different IND pre-training objectives, and then performs OOD clustering with the K-means clustering algorithm. In this paper, we employ two different pre-training objectives: CE+SCL and CE+KCL.
    
    \item \textbf{DeepCluster} This is an iterative clustering method proposed by \citet{Caron2018DeepCF}. In each iteration, firstly, K-means is used to assign pseudo labels to all unlabeled samples, and then the cross-entropy objective is used for representation learning. Due to the randomness of the clustering index, the cluster header parameters need to be reinitialized during each iteration. In the semi-supervised setting, we use the same IND pre-training objective as \citet{Zhang2021DiscoveringNI}

    \item \textbf{CDAC+} This is the first work of new intent discovery \cite{Lin2020DiscoveringNI}, and also the first work to propose a two-stage framework for clustering new intents in a semi-supervised setting. Firstly, it pre-trains a BERT-based \cite{devlin-etal-2019-bert} in-domain intent classifier then uses intent representations to calculate the similarity of OOD intent pairs as weak supervised signals.
    
    \item \textbf{DeepAligned} This is the second work of new intent discovery \cite{Zhang2021DiscoveringNI}. It is an advanced version of DeepCluster. The overall process of this method is basically the same as DeepCluster, and the only difference is that it designed a pseudo label alignment strategy to produce aligned cluster assignments for better representation learning. The method first performs K-means cluster assignments, and then performs representation learning. The two processes are iteratively performed in a pipeline manner, which results in the representation and cluster assignments not being updated simultaneously, leading to a suboptimal result.
    
    \item \textbf{DKT} This is the current state-of-the-art method for OOD intent discovery \cite{mou-etal-2022-disentangled}. In the IND pre-training stage, the CE and SCL objective functions are jointly optimized, and in the OOD clustering stage, instance-level CL and cluster-level CL objectives are used to jointly learn representation and cluster assignment. The main motivation is to design a unified multi-head contrastive learning framework to match the IND pre-training objectives and the OOD clustering objectives.

\end{itemize}

\section{Implementation Details}
\label{details}

For a fair comparison with previous work, similar with \citet{mou-etal-2022-disentangled}, we use the pre-trained BERT model (bert-base-uncased \footnote{https://github.com/google-research/bert}, with 12-layer transformer) as our network backbone, and add a pooling layer to get intent representation(dimension=768). Moreover, we freeze all but the last transformer layer parameters to achieve better performance with BERT backbone, and speed up the training procedure as suggested in \citet{Zhang2021DiscoveringNI}. We use two separate two-layer non-linear MLPs (ReLU as activation function) for instance-level head and cluster-level head. For the instance-level head, the output dimensionality is set to 128, and for the cluster-level head, the output dimensionality is set to the number of clusters.

In the IND pre-training stage, the training batch size is 128 and the learning rate is 5e-5; in the OOD clustering stage, the training batch size is 400 for Banking-10\%, Banking-20\%, Banking-30\%, HWU64-30\% and 512 for CLINC-30\% and the learning rate is 0.0003. Similar with \citet{mou-etal-2022-disentangled}, we use dropout \cite{Gao2021SimCSESC} to construct augmented examples for contrastive learning in OOD clustering stage with dropout rate 0.1. The temperatures of KCL and KCC are 0.5, and the cluster-level temperature is 1.0.
The augmented view is used as a new data point to participate in the KNN search process along with the original view. For the KCL objective function, in order to select K-nearest neighbors in a large enough search space and avoid using an excessively large batch size, we design an efficient dynamic queue mechanism. Specifically, in each iteration, for each sample in the batch, we randomly select 10 samples of the same type as it from the training set, and the queue length is maintained at 10*batch size. The queue length is set to 1280 in our implementation, which is the maximum value that the current device can bear. For the KCC method, the augmented view is used as a new data point to participate in the KNN search process along with the original view, and we set the threshold to 0.7 and K value to 400, which has been discussed in section \ref{hyper}.

We use Adam optimizer \cite{Kingma2015AdamAM} to train our model, and use the SC value of OOD data in the validation set as the basis for selecting the best checkpoints. All experiments use a single Tesla T4 GPU(16 GB of memory). 
The pre-training stage of our model lasts about 1 minute per epoch and clustering runs for 0.11 minutes per epoch on Banking-20\% \footnote{DKT almost consumes 0.5 minutes per epoch for pre-training and 0.08 minutes per epoch for clustering.}. The average value of the trainable model parameters is 17.34M, which is basically the same as DKT. It can be seen that our KCOD method has significantly improved performance compared to DKT, but the cost of time and space complexity is not large.

\section{Silhouette Coefficient (SC)}
\label{sc}
Following \citet{Zhang2021DiscoveringNI}, we use the cluster validity index (CVI) to evaluate the quality of clusters obtained during each training epoch after clustering. Specifically, we adopt an unsupervised metric Silhouette Coefficient \cite{Rousseeuw1987SilhouettesAG} for evaluation:
\begin{align}
    S C=\frac{1}{N} \sum_{i=1}^{N} \frac{b\left(\boldsymbol{I}_{i}\right)-a\left(\boldsymbol{I}_{i}\right)}{\max \left\{a\left(\boldsymbol{I}_{i}\right), b\left(\boldsymbol{I}_{i}\right)\right\}}
\end{align}
where $a\left(\boldsymbol{I}_{i}\right)$ is the average distance between $\boldsymbol{I}_{i}$ and all other samples in the $i$-th cluster, which indicates the intra-class compactness. $b\left(\boldsymbol{I}_{i}\right)$ is the smallest distance between $\boldsymbol{I}_{i}$ and all samples not in the $i$-th cluster, which indicates the inter-class separation. The range of SC is between -1 and 1, and the higher score means the better clustering results.

\section{Estimate Cluster C}
\label{c}
Since we may not know the exact number of OOD clusters, we use the following estimation method \cite{Zhang2021DiscoveringNI} to determine the number of clusters K before clustering. The method estimates C with the aid of the well-initialized intent features. We assign a big $K^{\prime}$ as the number of clusters at first. As a good feature initialization is helpful for partition-based methods (e.g., k-means), we use the pre-trained model to extract intent features. Then, we perform k-means with the extracted features. We suppose that real clusters tend to be dense even with $K^{\prime}$, and the size of more confident clusters is larger than some threshold $t$. Therefore, we drop the low confidence cluster whose size is smaller than $t$, and calculate K with:
\begin{align}
    K=\sum_{i=1}^{K^{\prime}} \delta\left(\left|S_{i}\right|>=t\right)
\end{align}
where $\left|S_{i}\right|$ is the size of the $i^{t h}$ produced cluster, and $\delta(\cdot)$ is an indicator function. It outputs 1 if the condition is satisfied, and outputs 0 if not. Notably, we assign the threshold $t$ as the expected cluster mean size $\frac{N}{K^{\prime}}$ in this formula.


\end{document}